\documentclass[letterpaper, 10 pt, conference]{ieeeconf}  

\IEEEoverridecommandlockouts                              

\usepackage{graphicx}
\usepackage{multirow}
\usepackage{amsmath}
\usepackage{booktabs}
\usepackage{comment}
\usepackage{flushend}

\makeatletter
\let\NAT@parse\undefined
\makeatother

\usepackage[breaklinks,colorlinks,bookmarks=false]{hyperref}

\usepackage[capitalize]{cleveref}
\crefname{section}{Section}{Sections}
\Crefname{section}{Section}{Sections}
\crefname{table}{Table}{Tables
\Crefname{table}{Table}{Tables}}
\crefname{figure}{Fig.}{Figs.}
\Crefname{figure}{Figure}{Figures}

\title{\LARGE \bf
RadarLCD: Learnable Radar-based Loop Closure Detection Pipeline
}

\author{\authorblockN{Mirko Usuelli, Matteo Frosi, Paolo Cudrano, Simone Mentasti and Matteo Matteucci}
\authorblockA{
Department of Electronics Information and Bioengineering\\
Politecnico di Milano\\
Milan, Italy\\
{\tt\footnotesize \{name.surname\}@polimi.it}
}
}

\begin{document}

\maketitle
\thispagestyle{empty}
\pagestyle{empty}

\begin{abstract}

Loop Closure Detection (LCD) is an essential task in robotics and computer vision, serving as a fundamental component for various applications across diverse domains. These applications encompass object recognition, image retrieval, and video analysis. LCD consists in identifying whether a robot has returned to a previously visited location, referred to as a loop, and then estimating the related roto-translation with respect to the analyzed location. Despite the numerous advantages of radar sensors, such as their ability to operate under diverse weather conditions and provide a wider range of view compared to other commonly used sensors (e.g., cameras or LiDARs), integrating radar data remains an arduous task due to intrinsic noise and distortion. To address this challenge, this research introduces RadarLCD, a novel supervised deep learning pipeline specifically designed for Loop Closure Detection using the FMCW Radar (Frequency Modulated Continuous Wave) sensor. RadarLCD, a learning-based LCD methodology explicitly designed for radar systems, makes a significant contribution by leveraging the pre-trained HERO (Hybrid Estimation Radar Odometry) model. Being originally developed for radar odometry, HERO's features are used to select key points crucial for LCD tasks. The methodology undergoes evaluation across a variety of FMCW Radar dataset scenes, and it is compared to state-of-the-art systems such as Scan Context for Place Recognition and ICP for Loop Closure. The results demonstrate that RadarLCD surpasses the alternatives in multiple aspects of Loop Closure Detection.

\end{abstract}


\section{INTRODUCTION}

In the field of robotics and computer vision, Loop Closure Detection (LCD) encompasses the integration of two sequential tasks. First, Loop Detection actively involves the identification of whether the moving robot has already visited the current location. Second, in case of a positive detection, Loop Closure estimates the roto-translation between the robot's poses corresponding to the two similar locations (i.e., the current and previously visited ones). 

One of the main uses of Loop Closure Detection is Simultaneous Localization and Mapping (SLAM), where the objective is to estimate the robot's poses while simultaneously constructing the map of the environment surrounding the robot. Nowadays, SLAM systems typically model the trajectory of the robot as a graph, the pose graph, hence the name Graph SLAM~\cite{grisetti2010tutorial, frosi2022art}. Moreover, SLAM consists of two key elements: the Tracker, responsible for short-term data associations, and the Loop Closure Detector, which identifies long-term data associations, known as loops. The Loop Closure Detector performs Loop Detection and Loop Closure by estimating the roto-translations between pairs of non-consecutive poses. While SLAM systems have evolved to incorporate LiDARs and cameras, integrating radar data remains a challenge, despite the many advantages, such as the ability to operate in various weather conditions~\cite{gao2021perception, hong2020radarslam}.

Loop Closure Detection has applications beyond SLAM. This approach has found significant use in various applications such as object recognition, image retrieval, and video analysis, demonstrating its broad impact on computer vision and related fields. For instance, in object recognition, LCD can detect and identify objects in images or video sequences.
Likewise, in image retrieval, this technique can help obtain images similar to a given query image by identifying similar loops in their feature trajectories.
Lastly, in video analysis, Loop Closure Detection can help detect repeated patterns or events in a video sequence by identifying loops in their feature trajectories, allowing the analysis of traffic or the detection of anomalies. 

Unlike other sensors, radars have attracted substantial interest owing to their exceptional ability to operate in challenging weather conditions and low-light environments. Indeed, Frequency Modulated Continuous Wave (FMCW) Radar sensors rely on radio waves, which are more robust to adverse weather conditions compared to laser beams (of LiDARs) because of their longer wavelength (0.8--10.0 cm) and the relative size of raindrops, fog, and snowflakes. Radars provide high-resolution information about surrounding objects and their movements, in the form of either polar or Cartesian images (useful for computer vision), making them valuable in various domains, including self-driving cars, surveillance, and robotics.

The current state-of-the-art brings attention to the limited research conducted on the application of FMCW Radar sensors for Loop Closure Detection. The current body of literature implies that further investigation into the application of radar sensors in LCD could be beneficial. This research could potentially bridge the existing gap in radar-related studies and contribute to advancements in SLAM research. In fact, existing methods in the radar field primarily focus on SLAM, particularly on tracking for odometry estimation~\cite{burnett_rss21, burnett2021we, utr, hong2020radarslam, 9197231, pou2021norm, 8460687, barnes2019masking, 8917111, 8794014, cfear} and place recognition~\cite{kim2020mulran, kim2018scan, gadd2020look, s20216002, tim2006radar, suaftescu2020kidnapped, utr}, posing challenges in integrating radar data into LCD methodologies. It is also important to distinguish between Place Recognition and Loop Closure Detection: the first retrieves a set of similar measurements to a query image offline, while the latter detects and estimates a loop roto-translation online.

This study presents RadarLCD, a novel approach to address the challenges of detecting and estimating loop roto-translation using the FMCW Radar sensor.
In our proposed model, we leverage the neural guidance provided by the radar-based tracker model HERO~\cite{burnett_rss21} to generate a 2D guidance map for key points discovery. These key points, along with their associated neural local descriptors, are utilized to construct a global descriptor for loop detection. The alignment method further facilitates effective key point selection during loop closure, thereby aiming to enhance the accuracy and reliability of SLAM systems that utilize FMCW Radar sensors. The contributions of this research can be summarized as follows.
\begin{enumerate}
\item We address the limitations of traditional sensor modalities, making RadarLCD scalable and robust in challenging environments, such as adverse weather conditions.
\item RadarLCD employs a supervised deep learning model for Loop Closure Detection. This choice is motivated by the radar sensor's susceptibility to distortion phenomena, enabling improved performance in detecting and estimating loop closures.
\item RadarLCD achieves generalization over never-seen scenes placed in other parts of the world with respect to the training dataset. This capability enhances the applicability and adaptability of the proposed approach to diverse real-world scenarios.
\item We compare RadarLCD with other methods for LCD, by performing an extensive evaluation over datasets differing by radar sensor used, i.e., the Radar RobotCar Oxford Dataset~\cite{RobotCarDatasetIJRR, RadarRobotCarDatasetICRA2020} and MulRan Dataset~\cite{kim2020mulran}, proving that the proposed method generalizes well on previously unseen data.
\end{enumerate}

These contributions collectively establish RadarLCD as a novel and effective solution for scalable and robust Loop Closure Detection using FMCW Radar sensor data.


\section{RELATED WORKS}\label{sec:related_works}

\paragraph{Camera}
One of the first learning-based works in Loop Detection was made by Xia et al.~\cite{xia2017evaluation} through vision. They asserted relevant achievements in fine-tuning common deep learning feature extractors, followed by a Support Vector Machine for binary classification.

Later, Wang Y. et al.~\cite{wang2022auv} presented a visual-based method for underwater Loop Closure Detection that employs a variational autoencoder network. This approach differs from traditional geometric methods and can effectively handle the dynamic and challenging underwater environment. Moreover, this method is unsupervised, which means that it does not require expensive labeled data, making it an efficient solution for underwater applications.

\paragraph{LiDAR}
Within the LiDAR-based literature, researchers have developed OverlapNet~\cite{chen2021overlapnet} to address the tasks of Loop Closure Detection using LiDAR-only sensor data. To achieve this, the authors employ a Siamese network as a feature extractor, which processes two LiDAR scans corresponding to distinct locations. These measurements are projected into multi-cue images, serving as input for the model. The extracted features are then utilized to classify loops as binary values based on the overlap between the two input sets. Additionally, the model estimates the yaw orientation by employing a 360-class cross-entropy loss function.

LCDNet~\cite{cattaneo2022lcdnet}, inspired by OverlapNet, is the current leading method for learning-based LiDAR LCD. Its main approach is based on 3D-spatial voxelized convolutions of classical CNNs. It uses a differentiable feature extractor to obtain local features, which are then fed into a NetVLAD pooling layer~\cite{arandjelovic2016netvlad}, also known as a global descriptor identifier.
Finally, the end-to-end pipeline is completed with a differentiable version of Singular Value Decomposition (SVD)~\cite{kleima1980singular, wang2021robust} to efficiently estimate a roto-translation.

Recent developments in attention-based architectures, such as Transformers~\cite{vaswani2017attention} for natural language processing and Vision Transformers~\cite{alexey2020vit, wu2020visual} for computer vision tasks, have motivated the use of attention mechanisms in point cloud registration. Building upon this trend, the most recent evolution of LCDNet is PADLoC~\cite{arce2023padloc}, which is a LiDAR-based Loop Closure Detection module that incorporates panoptic attention. PADLoC is a novel transformer-based head designed for point cloud matching registration, and it utilizes panoptic information during training to improve the model's robustness in registration tasks.

\begin{figure*}[!t]
    \centering
    \includegraphics[width=\textwidth]{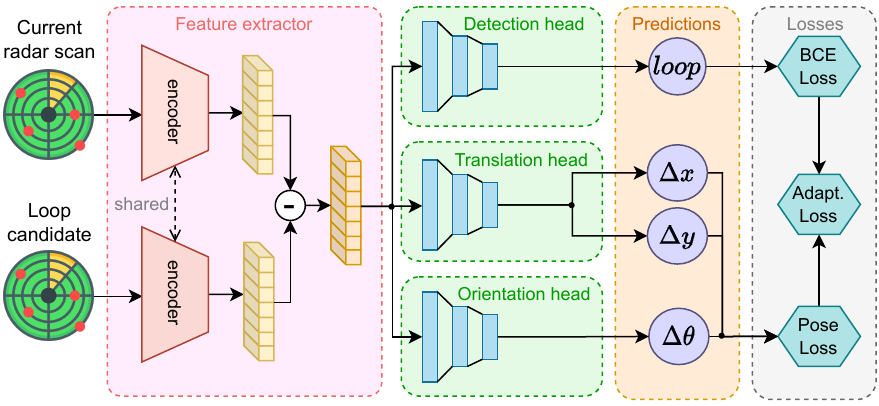}
    \caption{Architecture of the baseline end-to-end DMTL model. The complete pipeline is utilized during training, while at inference time, the model structure remains unchanged, except for the loss functions block. The encoder, which serves as the feature extractor, can vary between simple naive CNN, GoogleNet~\cite{googlenet}, ViT~\cite{wu2020visual, alexey2020vit}, and ConvNeXt~\cite{liu2022convnet}.}
    \label{fig:dmtl_model}
\end{figure*}

\paragraph{Radar}
Hong et al.~\cite{hong2020radarslam} developed a Loop Closure Detection method using radar images converted into 2D planar point clouds. They utilized a rotation-invariant global descriptor called M2DP~\cite{he2016m2dp} to represent the point cloud, which computes the density signature on the plane and extracts the left and right singular vectors of these signatures as the descriptor.

Later, Wang Z. et al.~\cite{wang2022train} fused convolutional neural network (CNN) features and line features to achieve accurate key location detection for Millimiter-Wave Radar in the loop closure detection part. It is worth noting that they utilize Millimeter-Wave Radar for short-range applications with high resolution, which differs from FMCW Radar as the latter is designed for long-range applications characterized by high accuracy and sensitivity.

Lastly, the recent TBV Radar SLAM~\cite{adolfsson2023tbv}, based on FMCW Radar, combines tightly coupled place similarity and odometry uncertainty search, loop descriptor creation from origin-shifted scans, and delayed loop selection until after verification to achieve a high correct-loop-retrieval rate. This method is designed to be robust against false constraints by carefully verifying and selecting the most likely loop constraints after registration. The authors build on the place recognition method Scan Context by Kim et al.~\cite{kim2018scan, kim2020mulran}, which detects loops and relative orientation by matching descriptors associated with multiple point clouds.

Radar-based place recognition has been the subject of several studies. S\v{a}ftescu et al.~\cite{suaftescu2020kidnapped} introduced a method that learns rotationally invariant descriptors within a metric feature space. Gadd et al.~\cite{gadd2020look} improved upon this approach by incorporating sequence-based place recognition using SeqSLAM~\cite{milford2012seqslam}. De Martini et al.~\cite{s20216002} proposed a two-stage system that integrates topological localization with metric pose estimation. Other research areas in perception include occupancy~\cite{8793263}, traversability~\cite{ai1040033, 9294415}, and semantic segmentation~\cite{9304674}. 

The "Under the Radar" framework~\cite{utr} presents a self-supervised solution designed for place recognition using radar-based odometry estimation and metric localization. It leverages a differentiable point-based motion estimator to learn robust key points, including their locations, scores, and descriptors, solely from localization errors.


\section{Proposed approach}
\label{sec:proposed_approach}

In order to address the current gap in the existing body of literature regarding the utilization of 2D radar scans, we initially introduce a fully deep learning model. This model is adapted from the cutting-edge work in the 3D-LiDAR field, specifically, OverlapNet as outlined in Chen et al.'s work~\cite{chen2021overlapnet}, detailed in \cref{sec:baseline} for reference as the baseline approach. This adaptation is crucial for facilitating a valid comparison with our novel contribution, RadarLCD, as presented in \cref{sec:RadarLCD}. A learning-based baseline has been preferred to be employed for this comparison as an assessment, since alternative 3D-LiDAR LCD methods would not yield significant insights due to dimensional disparity, except for Scan Context, which operates with 2D point cloud projections.
Furthermore, the decision to utilize a Deep Learning architecture was made to leverage the recurring patterns in radar scans, aiming to achieve dependable generalization capabilities. Therefore, a comparison with another neural-based work is essential to accurately evaluate RadarLCD as a novel advancement in the literature.

\subsection{Baseline model}
\label{sec:baseline}

Starting with a baseline model to serve as a reference in the FMCW Radar literature, we conducted a re-adaptation of the OverlapNet~\cite{chen2021overlapnet} architecture for radar scans. Expanding on the groundwork laid by OverlapNet, we have enhanced the initially designed DMTL (Deep Multi-Task Learning) architecture by specifically focusing on the task heads to also include translation prediction. Furthermore, we have incorporated a Binary Cross-Entropy loss function to enhance the accuracy of Loop Detection estimation.

This baseline architecture, depicted in \cref{fig:dmtl_model}, operates in an end-to-end manner following the Siamese paradigm. It accepts two radar scans as input, which are preprocessed to form single-channel Cartesian images. These images are then fed into a Siamese feature extractor, generating shared latent space embeddings. The delta operation, achieved through element-wise subtraction, yields a distinct input for the subsequent sub-task heads. We conducted experiments 
with different choices of encoder, considering GoogleNet~\cite{googlenet}, ViT~\cite{wu2020visual, alexey2020vit}, and ConvNeXt~\cite{liu2022convnet}. We also consider a CNN architecture composed of four convolutional layers and omitting max-pooling to preserve roto-translation patterns.

The architecture integrates dedicated sub-task heads for both Loop Closure and Loop Detection. 
The loop detection head is responsible for identifying loops, while the orientation head and translation head are involved in estimating rotation and translation, respectively. It is important to emphasize that while loop detection is performed immediately, the estimated loop closure is utilized at inference time as an initial guess for the ICP algorithm, as described by Cattaneo et al ~\cite{cattaneo2022lcdnet}. Each task head is associated with its own distinct loss function, which collectively contributes to the overall DMTL total loss function:

\begin{equation}
\mathcal{L}_{tot} = \alpha \cdot \mathcal{L}_{detection} + \beta \cdot \mathcal{L}_{closure}.
\end{equation}

\paragraph{Loop Detection head}
Using the embedding representation obtained from the subtraction operation as input, the detection head produces a binary classification score. This detection head is trained in a supervised manner, utilizing the Binary Cross-Entropy loss function:

\begin{equation}
\mathcal{L}_{detection} = -\frac{1}{N}\sum_{i=1}^{N}\left(y_i\log(p_i) + (1-y_i)\log(1-p_i)\right),
\end{equation}
where \(N\) is the total number of samples, \(y_i\) is the true label (either 0 or 1) for the \(i\)-th sample, and \(p_i\) is the predicted probability for the positive class for the \(i\)-th sample.

\paragraph{Loop Closure head}
The Loop Closure task is further divided into two sub-tasks: the translation head and the orientation head. The translation head is responsible for predicting the translation error, measured in meters, between the second input and the first input. On the other hand, the orientation head estimates the yaw angle between the two scans. As we operate in a 2D environment with FMCW Radar, the yaw angle indicates the orientation relative to the Z-axis, which is perpendicular to the radar scanning plane.

During the training process, we utilize a triplet association strategy that involves alternating between positive and negative sample pairs. It is important to emphasize that for negative pairs, we do not apply back-propagation to the loop closure head. This decision is made to preserve the regression capabilities that are specifically tailored to the Loop Closure task, differently from detection. Estimating a roto-translation for instances that are not loops would result in less meaningful results and could potentially degrade performance, especially when the instances are distant.

The proposed loss function includes the prediction of the $x$ and $y$ components of the translation using Euclidean distance, as well as the estimation of the angle $\theta$ in absolute value, expressed in radians for gradient magnitude considerations:
\begin{equation}
\begin{aligned}
\mathcal{L}_{closure} = {} & \lambda_{xy}\left(\frac{1}{N}\sum_{i=1}^{N}\sqrt{(x_i - \hat{x}_i)^2 + (y_i - \hat{y}_i)^2}\right)\\
& + \lambda_{\theta}\left(\frac{1}{N}\sum_{i=1}^{N}\left|\theta_i - \hat{\theta}_i\right|\right),
\end{aligned}
\end{equation}
where $\lambda_{xy}$ and $\lambda_{\theta}$ are weight parameters to control the importance of the distance and angle terms, respectively. $N$ is the total number of samples, $x_i$ and $y_i$ are the true x and y coordinates, $\hat{x}_i$ and $\hat{y}_i$ are the predicted x and y coordinates, $\theta_i$ is the true angle, and $\hat{\theta}_i$ is the predicted angle.

\subsection{RadarLCD}
\label{sec:RadarLCD}
RadarLCD draws inspiration from two sources: LCDNet~\cite{cattaneo2022lcdnet} and HERO~\cite{burnett_rss21}. The latter is an FMCW Radar tracker that utilizes unsupervised learning techniques to estimate consecutive roto-translations. It achieves this by selecting key points through a U-Net style architecture~\cite{ronneberger2015u}, which also generates guidance for key point suggestions. On the other hand, the LCDNet paradigm suggests employing triplet loss learning and using NetVLAD as a global descriptor generator for identifiers. By incorporating elements from both LCDNet and HERO, our proposal benefits from the strengths and insights of these previous works.

\begin{figure*}[th!]
    \centering
    \includegraphics[width=\textwidth]{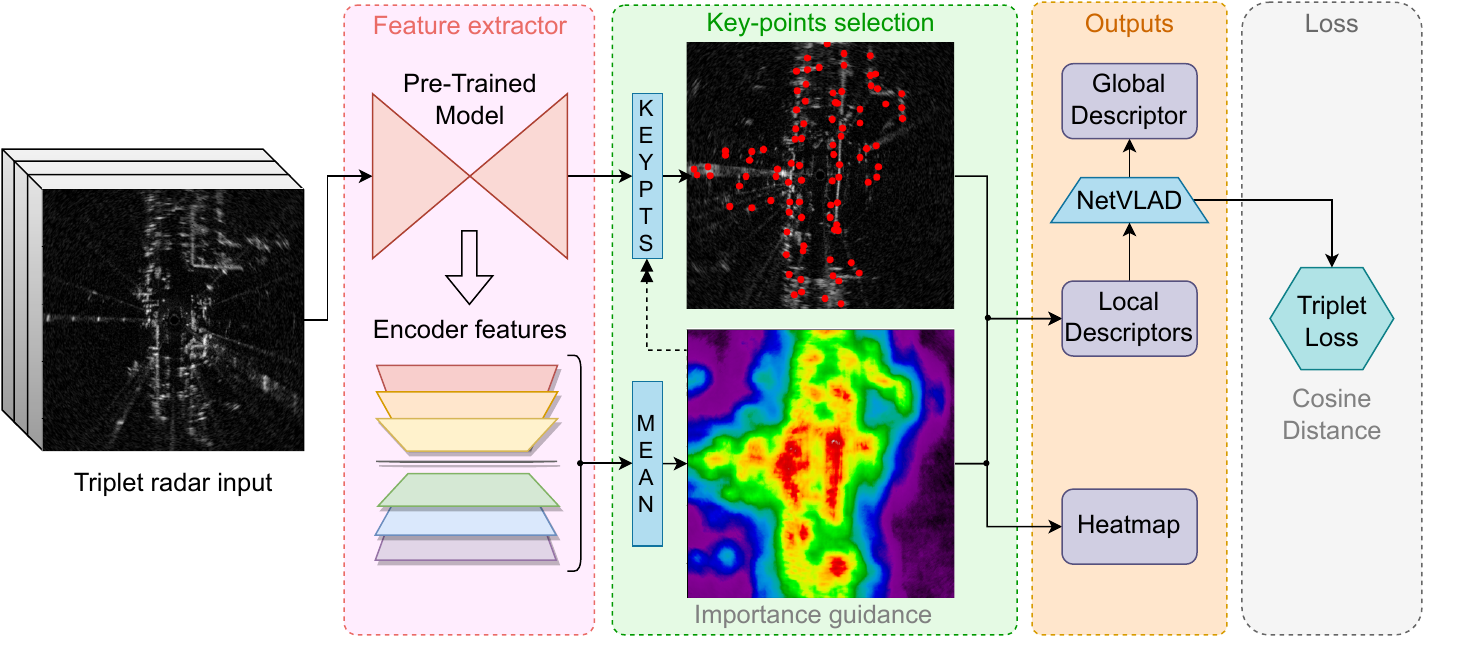}
    \caption{Training pipeline of RadarLCD model, while at inference time, the model structure remains consistent, with the exception of the loss block.}
    \label{fig:feature_based}
\end{figure*}

\Cref{fig:feature_based} illustrates the adaptation of the pre-trained U-Net model from HERO as the primary feature extractor. To make it suitable for the LCD task, we introduced a variation to the key points selection behavior. 
Differently from the standard approach, which was initially designed for tracking, we extract layer features from the U-Net encoder and compute a pixel-wise average across the feature channels to achieve channel-wise reduction. This operation generates a heat map that provides a 2D robust guidance for the selection of key points.

We initialize these with a uniform distribution in the 2D space and iteratively refine their positions based on the guidance. The corresponding module utilizes the received guidance to improve the convergence of the key points. This iterative process ensures that more meaningful key points are obtained. Furthermore, we extract local descriptors pixel-wise at the locations of the key points within the U-Net features. These local descriptors are then used in conjunction with the key points for subsequent analysis and processing as one of the model outputs along with the heatmap which can be also exploited as a filtering mask.

The pipeline methodology described here draws some parallels to the approach taken by "Under the Radar" in the context of place recognition. However, there are two key differences in our pipeline. Firstly, while "Under the Radar" focuses on offline location retrieval, our goal is centered around online detection.
Secondly, our pipeline diverges in terms of how we utilize the descriptors. For each radar scan, they construct a dense descriptor map that facilitates the key point matching discussed earlier. While these descriptors were originally trained for pose estimation, they re-use them to generate a location-specific embedding. 
In contrast, we employ a channel-wise average pooling technique to reduce all the features of the encoder. This process generates a smoother guidance map, with distinct peaks corresponding to regions of interest, enabling the detection of significant features in the scan.

In line with the methodology of LCDNet~\cite{cattaneo2022lcdnet}, the selected local descriptors corresponding to the identified key points are fed into a NetVLAD pooling layer~\cite{arandjelovic2016netvlad}, which generates the final global descriptor for the Loop Detection task. Following the suggestion of Arandjelović et al.~\cite{arandjelovic2016netvlad}, the training of NetVLAD is configured using the triplet loss function, ensuring that the model achieves optimal performance and exhibits discriminative capabilities.

By employing the triplet loss function, the model is encouraged to learn embeddings that effectively discriminate between loops and non-loops pairs of radar images:

\begin{equation}
    \mathcal{L}_{triplet} = \max(0, d(a, p) - d(a, n) + \delta),
\end{equation}
where $d(a, p)$ is the distance between the anchor $a$ and the positive $p$, $d(a, n)$ is the distance between the anchor $a$ and the negative $n$, and the $\delta$ is a hyper-parameter that controls the minimum difference between the two distances. 

In the proposed pipeline, the cosine distance is adopted as the distance metric:
\begin{equation}
    d(\mathbf{u}, \mathbf{v}) = CosineDistance(\mathbf{u}, \mathbf{v}) = 1 - \frac{{\mathbf{u} \cdot \mathbf{v}}}{{\|\mathbf{u}\| \cdot \|\mathbf{v}\|}},
\end{equation}
where $\mathbf{u}$ and $\mathbf{v}$ are the two global descriptors being compared, $\lVert \mathbf{u} \rVert$ and $\lVert \mathbf{v} \rVert$ are the magnitudes of the vectors.

At inference time, the cosine distance is employed to compare global descriptors that represent radar images and determine whether a loop has occurred. The cosine distance is preferred over the Euclidean distance for several reasons. One primary advantage is that it takes into account both the magnitude and orientation of the compared vectors. In many applications, the orientation or direction of vectors is often more significant than their length or magnitude. Additionally, as in this case, the cosine distance is particularly useful for high-dimensional data, where the Euclidean distance can be less effective due to the "curse of dimensionality", leading to increased sparsity and diminished meaningfulness of distances between points. In contrast, the cosine distance focuses on the angles between vectors rather than their absolute distances, making it a suitable metric.

\section{Experiments}
We conducted a comprehensive comparison across the Radar RobotCar Oxford Dataset~\cite{RobotCarDatasetIJRR, RadarRobotCarDatasetICRA2020} and MulRan Dataset~\cite{kim2020mulran}, as mentioned earlier. For each scene within these datasets, a dedicated testing set was created, taking into consideration the GNSS position associated with each measurement. The definition of a loop followed the approach outlined by Cattaneo et al.~\cite{cattaneo2022lcdnet}, as described in \cref{sec:proposed_approach}, and it was established with a maximum distance of 4 meters. It is worth noting that every scene in the datasets was balanced for binary classification, ensuring that performance evaluation is not affected by any imbalances.

In order to align with the pre-trained model of HERO, we trained both architectures on the majority of scenes from the Oxford dataset, as described by Burnett et al ~\cite{burnett_rss21}. The evaluation of these models was performed on the MulRan dataset, which includes scenes from DCC, KAIST, and Riverside, as well as the remaining Oxford scenes that were not included in the training set. However, we excluded the Sejong scene from the MulRan dataset due to the absence of loops. To ensure a robust performance comparison, benchmark datasets were created by averaging scenes of the same type, excluding any training data.

\subsection{Loop Detection evaluation}

It is essential to note that evaluating Loop Closure Detection requires a distinct approach and should not solely rely on Recall@N as the primary metric, as is done in the Place Recognition task. Instead, we rely on Mean Average Precision (mAP), which is particularly well suited for this task as it allows for the consideration of multiple thresholds, capturing variations in recall and providing a comprehensive assessment.
We considered a threshold spanning from 0.25 to 0.90 with a step of 0.05. This choice resulted in the utilization of 13 distinct thresholds, enabling a comprehensive comparison of the Loop Detection capabilities.

\begin{table*}[th!]
  \centering
  \caption{Mean Average Precision for Loop Detection}
  \begin{tabular}{@{}lcccc@{}}
    \toprule
    & Oxford & DCC & KAIST & Riverside \\ 
    Model & $mAP\ [\%]$ & $mAP\ [\%]$ & $mAP\ [\%]$ & $mAP\ [\%]$ \\
    \midrule
    LiDAR Scan Context \cite{kim2018scan} & 51.30 & 63.90 & 73.99 & 73.86 \\ 
    Radar Scan Context \cite{kim2020mulran} & 50.41 & 51.80 & 50.84 & 52.39 \\
    \midrule
    \textbf{RadarLCD}$^{\dag}$ & \textbf{94.27} & \textbf{94.20} & \textbf{94.03} & \textbf{90.17} \\
    Baseline CNN & 69.44 & 53.16 & 60.75 & 63.61 \\
    Baseline GoogleNet & 77.38 & 51.72 & 51.49 & 56.77 \\
    Baseline ViT & 72.70 & 47.87 & 53.13 & 58.74 \\
    Baseline ConvNeXt & 64.70 & 61.29 & 60.51 & 58.57 \\
    \bottomrule
  \end{tabular}
  \label{tab:loop_detection}
\end{table*}

\begin{table*}[th!]
  \centering
  \caption{Rotation error ($R_{\epsilon}$) and translation error ($T_{\epsilon}$) for Loop Closure}
  \begin{tabular}{@{}l cc cc cc cc@{}}
    \toprule
    & \multicolumn{2}{c}{Oxford} & \multicolumn{2}{c}{DCC} & \multicolumn{2}{c}{KAIST} & \multicolumn{2}{c}{Riverside} \\
    Model & $R_{\epsilon}\ [^\circ]$ & $T_{\epsilon}\ [m]$ & $R_{\epsilon}\ [^\circ]$ & $T_{\epsilon}\ [m]$ & $R_{\epsilon}\ [^\circ]$ & $T_{\epsilon}\ [m]$ & $R_{\epsilon}\ [^\circ]$ & $T_{\epsilon}\ [m]$\\
    \midrule
    Cen and Newman~\cite{cen2019radar} & \textbf{~2.38} & ~2.62 & ~2.55 & \textbf{~3.90} & ~~2.92 & \textbf{~3.94} & ~1.34 & ~3.18\\
    \midrule
    \textbf{RadarLCD}$^{\dag}$ & ~2.89 & \textbf{~2.53} & \textbf{~1.33} & ~4.04 & \textbf{~~2.40} & \textbf{~3.94} & \textbf{~0.76} & ~3.16\\
    Basline CNN & 12.62 & ~3.70 & 36.43 & ~4.06 & ~~41.03 & ~4.26 & 31.68 & ~3.85\\
    Baseline GoogleNet & ~8.76 & ~3.08 & ~2.14 & ~4.00 & ~~2.97 & ~4.08 & ~1.04 & ~3.46\\
    Baseline ViT & ~8.86 & ~3.21 & ~8.32 & ~4.01 & ~~8.90 & ~4.09 & ~3.74 & ~3.49\\
    Baseline ConvNeXt & ~9.98 & ~3.13 & ~2.95 & ~4.00 & ~~4.49 & ~4.06 & ~1.77 & ~3.47\\
    \bottomrule
  \end{tabular}
  \label{tab:loop_closure}
\end{table*}

\Cref{tab:loop_detection} presents a detailed comparison of our proposed method RadarLCD (\cref{sec:RadarLCD}) against state-of-the-art methods, consisting of LiDAR and radar variations of Scan Context~\cite{kim2018scan, kim2020mulran}. We compare also against our baseline model (\cref{sec:baseline}), which is fully-learned, using various encoders (CNN-like model, GoogleNet~\cite{googlenet}, ViT~\cite{wu2020visual, alexey2020vit}, and ConvNeXt~\cite{liu2022convnet}).
The results highlight the significant superiority of RadarLCD alternative over Scan Context and all variations of the baseline models. By employing different thresholds, RadarLCD demonstrates its effectiveness in accurately identifying loop occurrences with high sensitivity. Qualitatively, the rise in performance
observed can be attributed to Deep Learning’s capacity to
generalize from recurrent intrinsic patterns in radar scans,
in contrast to hand-crafted methods.

\subsection{Loop Closure evaluation}
To evaluate Loop Closure, our evaluation methodology concentrates on measuring the translation and orientation errors with respect to the GNSS ground truths. The translation error is computed using the L2 norm in the XY-plane, while the orientation error is determined using the L1 norm, as the absolute difference between the predicted and ground truth headings.

We compare against the SOTA radar keypoint finding method presented by Cen and Newman~\cite{cen2019radar}, as well as with all the fully-deep baselines seen in \cref{tab:loop_detection}.
Upon observing \cref{tab:loop_closure}, it becomes apparent that the initial guess provided by DMTL baselines does not consistently enhance Loop Closure when compared to performing ICP without any initial guess. However, the utilization of key points identified by RadarLCD model has shown improvement over the hand-crafted methods represented by Cen and Newman~\cite{cen2019radar} when ORB~\cite{orb} descriptors are applied to them for RANSAC matching. This highlights the effectiveness of leveraging the key points extracted by our proposal in enhancing the performance of Loop Closure compared to traditional hand-crafted approaches.

\section{Conclusion}

In this research, we proposed RadarLCD, a novel deep-based pipeline for Loop Closure Detection using FMCW Radar sensors, representing the first deep learning-based solution of its kind in the literature. Notably, our proposal surpasses the current state-of-the-art method, Scan Context, in terms of FMCW Radar Loop Detection. Moreover, by leveraging the more robust guidance provided by the feature extraction process performed by our model, we achieve superior key point selection compared to the current state-of-the-art in FMCW Radar. This highlights the substantial contribution of our approach in enhancing both Loop Closure Detection performance and the effectiveness of key point selection in the FMCW Radar domain.
We underline that the use of a pre-trained network on Radar Odometry leverages features already trained and enhances the modularity of our approach. This modularity is a novelty of our work and opens for further contributions.



\bibliographystyle{IEEEtran}
\bibliography{IEEEabrv, mybib}

\end{document}